\documentclass[twocolumn,9pt]{article} 

\usepackage[square,numbers,sort&compress,comma]{natbib}

\usepackage{amsmath}
\usepackage{amssymb}
\usepackage{caption}
\usepackage{graphicx}
\usepackage{latexsym}
\usepackage{times}
\usepackage[pagewise]{lineno}
\usepackage{hyperref}
\usepackage{algorithm}
\usepackage{algpseudocode}
\usepackage{xcolor}
\usepackage{eso-pic}
\topmargin - 12pt 
\renewenvironment{abstract}%
              {
               \small
               {\bfseries \abstractname}
               \par
               \vspace{10pt}
              }

\renewcommand\abstractname{Abstract}

\newcommand{\nomenclature}
              [1]
              {
               \bgroup
               \flushleft
               \small\bf
               #1
               \par
               \egroup
              }

\renewcommand{\section}
              [1]
              {
               \bgroup
               \flushleft
               \small\bf
               \refstepcounter{section}
               \arabic{section}. #1
               \par
               \egroup
              }

\renewcommand{\subsection}
              [1]
              {
               \bgroup
               \flushleft
               \small\em
               \refstepcounter{subsection}
               \arabic{section}.
               \arabic{subsection}. #1
               \par
               \egroup
              }

\renewcommand{\subsubsection}
              [1]
              {
               \bgroup
               \flushleft
               \small\em
               \refstepcounter{subsubsection}
               \arabic{section}.
               \arabic{subsection}.
               \arabic{subsubsection}. #1
               \par
               \egroup
              }

  \newcommand{\acknowledgement}
              [1]
              {
               \bgroup
               \flushleft
               \small\bf
               #1
               \par
               \egroup
              }

  \newcommand{\sectionbib}
              [1]
              {
               \bgroup
               \flushleft
               \small\bf
               #1
               \par
               \egroup
              }

\AddToShipoutPictureFG*{%
  \AtPageLowerLeft{%
    \hspace*{0.17\paperwidth}%
    \raisebox{2.5cm}{\small Accepted for publication in \emph{Proceedings of the Combustion Institute}.}%
  }%
}

\begin{document}



\small
\baselineskip 10pt

\setcounter{page}{1}
\title{\LARGE \bf An Approximate Graph Elicits Detonation Lattice}

\author{{\large Vansh Sharma$^{a,*}$ and Venkat Raman$^{a}$}\\[10pt]
        {\footnotesize \em $^a$Department of Aerospace Engineering, University of Michigan, Ann Arbor, MI 48109-2102, USA}\\[-5pt]
        }

\date{}  

\twocolumn[\begin{@twocolumnfalse}
\maketitle
\rule{\textwidth}{0.5pt}
\vspace{-5pt}

\begin{abstract} 
This study presents a novel algorithm based on graph theory for the precise segmentation and measurement of detonation cells from 3D pressure traces, termed detonation lattices, addressing the limitations of manual and primitive 2D edge detection methods prevalent in the field. Using a segmentation model, the proposed training-free algorithm is designed to accurately extract cellular patterns, a longstanding challenge in detonations research. First, the efficacy of segmentation phase on two synthetic datasets is evaluated with an error of 2\%. Next, 3D simulation data is used to establish performance of the graph-based workflow. The results of statistics and joint probability densities show oblong cells aligned with the wave propagation axis with 17\% deviation, whereas larger dispersion in volume reflects cubic amplification of linear variability. Although the framework is robust, it remains challenging to reliably segment and quantify highly complex cellular patterns. However, the graph-based formulation generalizes across diverse cellular geometries, positioning it as a practical tool for detonation analysis and a strong foundation for future extensions in triple-point collision studies. 
\end{abstract}

\vspace{10pt}

{\bf Novelty and significance statement}

\vspace{10pt}
Detonation cells are key features of canonical propagating waves, and are useful for understanding cellular instabilities. Currently, there is very limited work on methods for estimating cell sizes, with manual counting in two-dimensions being the common approach. Existing numerical approaches typically work with 2D data, require extensive manual intervention, or rely on 2D edge heuristics that lack robustness and applicability across 3D volumetric data. 
This work, to the best of the authors' knowledge, is the first to present a generalized algorithm that automatically detects, segments, and quantitatively measures three-dimensional detonation cells using graph theory. 
The proposed method provides geometric structure of three-dimensional cells, vastly improving the ability to understand detonation structures. These findings will ultimately help improve practical detonation applications (rotating, pulse, oblique and standing detonation wave engines).

\vspace{5pt}
\parbox{1.0\textwidth}{\footnotesize {\em Keywords:} 3D Detonation; Soot Foil; Graph; Cell Classification; Cellular Detonation; Detonation Lattice; SAM Model}
\rule{\textwidth}{0.5pt}
*Corresponding author.

\vspace{5pt}
\end{@twocolumnfalse}] 

\section{Introduction\label{sec:introduction}} \addvspace{10pt}
As detonation-based devices~\cite{nicholls1957intermittent} were pioneered in the early 1950s, soot foils emerged as a diagnostic tool to visualize \textcolor{black}{the two-dimensional cellular imprint of triple-point trajectories} and associated wave-collision points on the confining surface, to study intrinsic high-speed combustion and shock phenomena. These soot-coated substrates are placed along gaseous detonation channels that record triple-point tracks as the wave propagates, producing a ``cellular lattice'' imprint of the detonation front. Despite widespread use, these techniques are inherently two-dimensional, sampling only wall-adjacent dynamics and missing the full 3D front morphology. This motivates techniques to map the soot-foil lattice patterns to a 3D volumetric representation with minimal human effort.

\begin{figure*}[!ht]
    \centering
  \includegraphics[width=0.99\linewidth]{ 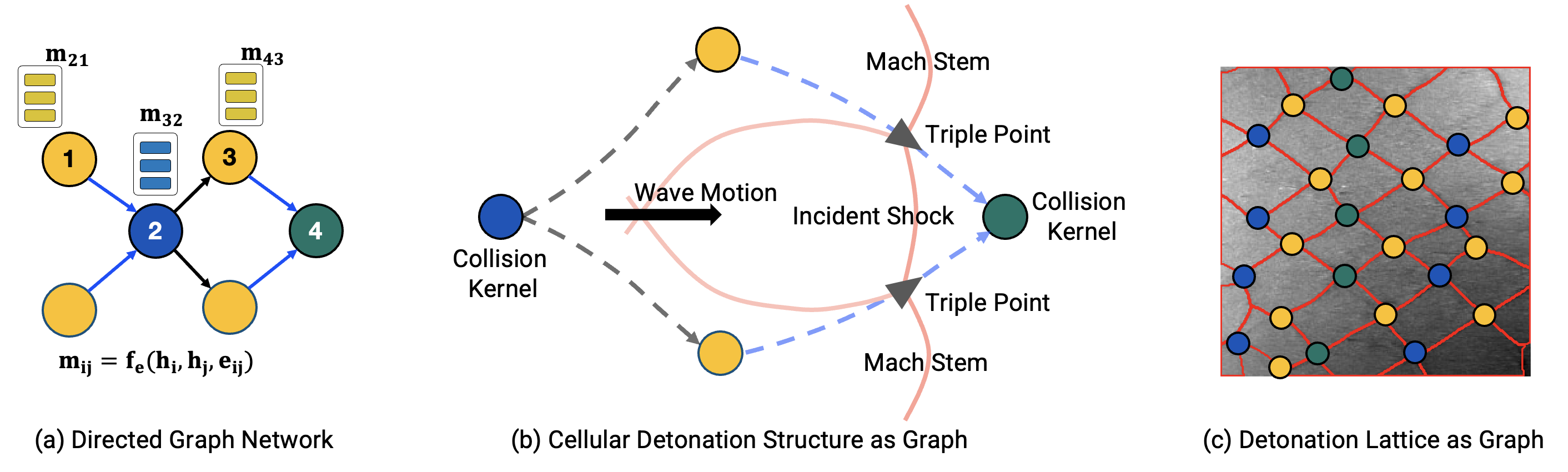}
  \caption{\footnotesize Relating graph message passing to cellular detonation structure: (a) Directed message-passing graph: node features $h_i$ exchange edge ($e_{ij}$) messages $m_{ij}$, (b) Cellular detonation interpreted as a graph: triple points and collision kernels serve as nodes connected along Mach stems and incident shocks; wave motion induces directional interactions and (c) 2D detonation lattice as a graph, with collisions points as nodes and wave path as edges, illustrating how collisions localize within the network.}
  \label{fig:detonationGraph}
\end{figure*}

Several techniques have been developed to quantify the cellular pattern on 2D soot foils~\cite{nair2023detonation, carter2022direct, shepherd1988analyses, ng2024detonation, jalontzki2025computer}. Shepherd et al.~\cite{shepherd2009detonation} proposed a PSD-based cell-size estimator coupled with edge detection, linking pattern periodicity to dominant spatial frequencies via 2D Fourier analysis. Subsequent workflows and software packages popularized this route but retained substantial manual tuning. Carter et al.~\cite{carter2022direct} extracted directional gradients with manual input to overlay binary cell maps while Ng et al.~\cite{ng2024detonation} performed CH*-based extraction of soot-foil-like patterns and then measured $\lambda$ (characteristic cell width) with tools derived from Shepherd et al.~\cite{shepherd_tool}. Siatkowski et al.~\cite{siatkowski2024predicting} showed CAD-assisted line plotting on foils and later trained model on curated databases to predict $\lambda$. Neural network approaches have since expanded, using larger datasets to predict $\lambda$ from image features, though segmentation of irregular lattices remains challenging. Although recent approaches in computer vision~\cite{jalontzki2025computer} leverage contour-based segmentation, these methods are brittle in the presence of practical irregularities and topological defects. In contrast, our previous study~\cite{sharma2025machine} established a generalizable machine learning (ML) framework based on cellular biology models and validated its efficacy with rigorous benchmarks that exceed baseline techniques~\cite{shepherd_tool, jalontzki2025computer}.

Across these methods, the lattice pattern is treated as a 2D tessellation without any depth, while true 3D morphology is closer to practical systems. Monnier et al.~\cite{monnier2022analysis} used soot-plate experiments with front-view and longitudinal recordings, showing longitudinal traces alone can misrepresent the full front; they proposed tessellation/Voronoi-based measures for an average cell width but note irreducible counting uncertainty and the reduction of 3D structure to 2D imprints. Later, Monnier et al.~\cite{monnier2023graph} used a graph-theoretic framework that infers a pseudo-3D cell width from 2D front-view detonation patterns, linking polygonal lattices to a volumetric scale without directly measuring the lattice itself. Borisov \& Kudryavtsev~\cite{borisov2017investigation} characterized 3D detonation cells using high-fidelity numerical simulations guided by linear instability theory: predicted transverse wavelengths were compared with simulation fields, cell evolution was quantified via Fourier analysis of transverse velocity, and triple-point paths were visualized using maximum-pressure isosurfaces (``spatial smoke foils"). Their main limitations were acknowledged resolution constraints for early/small cells and simplified chemistry. 
Although high-fidelity simulations can generate 3D volumetric cellular fields~\cite{williams1996detailed, crane2023three, iwata2023direct, tsuboi2002three, abisleiman2025thermochemical, abisleiman2025structure}, to our knowledge, there is no direct procedure that reconstructs or measures a 3D structure directly from detonation lattice data. The focus of this work is to bridge this gap by developing an algorithm that (i) robustly segments 3D lattice patterns, (ii) quantify anisotropy/regularity beyond $\lambda$, and (iii) is consistent with physics-based constraints. The algorithm is a generalizable, minimally supervised approach that captures 3D lattices from volumetric data. The following sections detail the algorithm for graph construction ($\S$\ref{sec:GraphCon}) followed by results of the lattice case ($\S$\ref{sec:results_toy}) and a 3D numerical detonation lattice ($\S$\ref{sec:results-3Dsim}). 


\section{Methodology\label{sec:method}} \addvspace{10pt}

Graph representations provide mathematically rigorous encoding of interaction topology in natural systems, enabling statistically robust inference and analysis~\cite{sanchez2020learning}.
A graph provides a physically meaningful surrogate for the underlying 3D detonation lattice: nodes correspond to triple point collision kernels, while edges represent the intervening segments (Mach stems, \textcolor{black}{triple point path} and incident shocks) oriented along wave motion (see Fig.~\ref{fig:detonationGraph}). This abstraction preserves the essential topology—adjacency, branching, and cycles—that governs cell formation, irrespective of geometric distortions in any particular view. A graph built from these elements concisely should encode the 3D organization of the front: local motifs (Y-junctions, loops) assemble into a global lattice whose statistics (degree, cycle length, path structure) map to cell size, anisotropy, and interaction frequency.

\subsection{Volumetric data segmentation\label{sec:volumedataSeg}} \addvspace{10pt}
Building on our prior work~\cite{sharma2025machine}, we implement a slice-wise volumetric segmentation workflow: the 3D field is decomposed into an ordered stack of 2D slices, each undergoes targeted preprocessing (denoising, contrast normalization, structure-enhancing filters) to suppress noise and amplify interfaces, and the preprocessed sequence, preserving slice order, is passed to a Segment-anything–style (SAM) model for inference~\cite{kirillov2023segment, pachitariu2025cellposeSAM}. 
SAM-based foundation models provide strong, reusable visual representations for segmentation across diverse image distributions \cite{kirillov2023segment,sharma2026autohood3d,ravi2024sam}. Cellpose-SAM repurposes the pretrained SAM image encoder but replaces sequential mask decoding with Cellpose’s dense flow-field formulation, enabling efficient instance segmentation and straightforward extension to 3D by averaging predictions across XY/ZY/YZ slices before performing 3D mask dynamics~\cite{pachitariu2025cellposeSAM}. While this architecture is attractive for automated cell-pattern extraction, at the same time, for strictly 2D detonation-cell post-processing (e.g., soot-plate or planar visualizations), deploying a large foundation model can be unnecessary overhead; models such as ``cyto3" pipelines~\cite{sharma2025machine} may be sufficient when the goal is a single cell-size metric rather than general-purpose segmentation. Also noted in~\cite{pachitariu2025cellposeSAM},  direct comparison between Cellpose3 (cyto3) and CellSAM on datasets both were trained on, \emph{``cyto3 performs better on every dataset"}. Thus, careful consideration is required before deploying such models: for 2D soot-foil analysis, lighter non-SAM pipelines are often sufficient and preferable due to their smaller memory footprint. In contrast, for volumetric segmentation tasks, SAM-based approaches can offer improved robustness and accuracy relative to the CNN-based models~\cite{sharma2025machine,pachitariu2025cellposeSAM}.

Following initial analysis, SAM-based design enables flexible, multi-resolution operation across volumes of varying size and voxel aspect ratio while preserving fidelity to the underlying 3D morphology. The model returns a set of masks that uniquely label each cell, assigning a distinct integer ID to every voxel within each node and across nodes when aggregated. \textcolor{black}{Each uniquely labeled cluster is then reduced to a single representative centroid, defined as the coordinate-wise average of its distance-transform-weighted (DTW) center and largest-inscribed-sphere (LIS) center~\cite{coxeter1973regular}. This definition is important because it suppresses bias from thin protrusions and irregular boundaries while maintaining a representative point within the true interior of the cluster.}


\subsection{Graph construction\label{sec:GraphCon}} \addvspace{10pt}

From $\S$\ref{sec:volumedataSeg}, given labeled 3D centroids $C\in\mathbb{R}^{N\times 3}$ with integer instance labels $L$ (using SAM model and LIS centers), a sparse undirected graph is constructed by proposing edges only \emph{forward} along a user-chosen axial direction $D\in\{\pm X,\pm Y,\pm Z\}$ (typically aligned with the wave propagation), followed by local pruning and degree capping. The detailed pseudo-code is presented in Algorithm~\ref{algo:detGraph}. Each centroid is discretized into rectilinear bin coordinates $(I,J,K)$ using user-defined grids $(u_x,u_y,u_z)$, and the direction $D$ induces an axial index $A\in\{I,J,K\}$, lateral indices $(U,V)$, and a sign $\mathrm{sgn}\in\{+1,-1\}$. 

Nodes are processed in lexicographic order consistent with $\mathrm{sgn}$; an optional reverse pass is included to reduce ordering bias. For a root node $i$, a candidate node $j$ is considered only if it satisfies: (i) strict axial advance $\mathrm{sgn}\,(A_j-A_i)>0$; (ii) bounded axial span $|A_j-A_i|\le A_{\max}$; and (iii) bounded lateral offset $|U_j-U_i|+|V_j-V_i|\le R_{\text{side}}$. \textcolor{black}{$A_{\max}$ and $R_{\mathrm{side}}$ define only the local axial and transverse search window for candidate edges and are selected (sufficiently large enough) relative to the bin/voxel resolution so as to capture plausible local neighbors while suppressing nonlocal jumps. The final retained connections are subsequently determined by the blocking and occupancy tests, as discussed next.} Remaining candidates are ranked by increasing axial separation and then lateral distance (with axial separation taken as either $|A_j-A_i|$ in bin units or $(C_j-C_i)\cdot \hat{d}$ in continuous units, depending on implementation), and the \textcolor{black}{top $K_{\mathrm{cand}}$} candidates are retained.

\begin{figure}[!ht]
    \centering
  \includegraphics[width=0.85\linewidth]{ 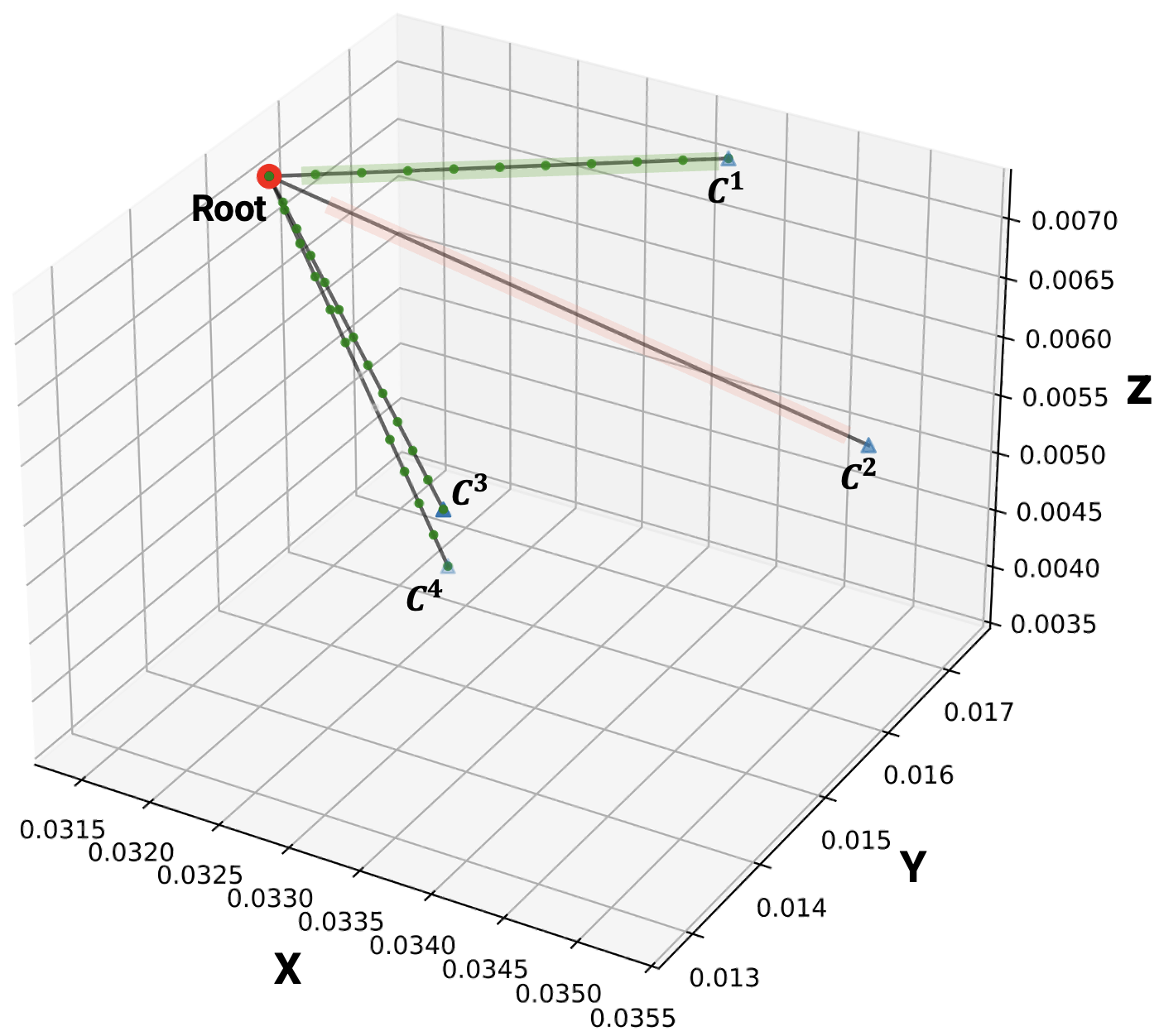}
  \caption{Search in Euclidean space: a root node proposes its top four beam-search candidates; after applying occupancy constraints, only the green-marked (physically consistent) candidates are accepted, while the red-marked is rejected.}
  \label{fig:beam}
\end{figure}

Edges are greedily accepted from the ranked shortlist subject to a per-node degree cap $\deg_{\max}$ and duplicate suppression (undirected edges are stored as $(\min(i,j),\max(i,j))$). Two optional geometric tests further filter edges: (a) a \emph{proximity} test (\textsc{Cluster}) that requires each endpoint centroid to lie within a threshold $\tau$ of voxels belonging to the other endpoint's label, implemented with per-label KD-trees; and (b) an \emph{occupancy} test (\textsc{Between}) that samples points along the segment $(C_i,C_j)$ at a step size proportional to the minimum grid spacing and requires a minimum hit fraction $\phi_{\min}$ of sampled locations to fall within a voxel radius $r_{\text{vox}}$ of any labeled voxel. 
In terms of physics, these constraints ensure that a wave must have traversed along/in the vicinity of the connecting path between nodes, ensuring that the edge is part of the original detonation lattice structure and suppress spurious long-range connections. Figure~\ref{fig:beam} illustrates the occupancy test for the multi-axis node search: from a given root centroid, a Euclidean neighborhood search identifies the top beam-search candidates $(C_1 \text{ to } C_4)$ in the forward direction. Proposed edges are then screened with an occupancy (between-voxels) constraint that checks whether the segment lies within labeled structure along its path. Candidates that satisfy this physical-consistency test are retained (green), while spurious connections lacking supporting occupancy are rejected (red). \textcolor{black}{We note that Fig.~\ref{fig:beam} serves only as a skeletal illustration for reconstruction; the cell volume is not computed under a bipyramidal or other prescribed analytic shape assumption, but directly from the reconstructed closed surface using its triangulated geometry.}
With the \textcolor{black}{top-$K_{\mathrm{cand}}$} shortlist and the degree cap, the resulting graph has $\mathcal{O}(N)$ edges, while KD-tree queries keep the optional test costs sublinear in the number of labeled voxels. 

\textcolor{black}{This will be demonstrated in the studies below, where the algorithm exhibits a clear separation between the search envelope and the acceptance parameters. $A_{\max}$ and $R_{\mathrm{side}}$ determine the axial and lateral extent of the candidate search, so increasing them enlarges the proposal set and computational cost, while decreasing them too far may exclude valid local neighbors. In contrast, the occupancy threshold $\phi_{\min}$ is the dominant physical-consistency control: higher values produce a more conservative graph, whereas lower values admit weaker connections. The candidate cutoff $K_{\mathrm{cand}}$ mainly affects recall prior to pruning, $deg_{\max}$ sets the final graph sparsity, and the blocking/skip parameters regularize topology by suppressing centroid-crossing and shortcut edges. In the present implementation, the occupancy sampling step and tube radius are scaled with the minimum grid spacing, which helps reduce sensitivity to absolute mesh resolution.}

\begin{algorithm}[!ht]
\caption{Lattice Graph Construction}
\begin{algorithmic}[1]
\Require $C\!\in\!\mathbb{R}^{N\times3}$, $L$; grids $u_x,u_y,u_z$; axis, $A_{\max}$, $R_{\text{side}}$, $K_{\mathrm{cand}}$, $\text{deg}_{\max}$; optional gates: \textsc{Cluster}$(\tau)$, \textsc{Between}$(s_{\text{vox}},r_{\text{vox}},\phi_{\min})$
\State Map axis $\to(\text{ax},\mathrm{sgn},u,v)$; bin: $I\!=\!\text{bin}(C_x,u_x)$, $J\!=\!\text{bin}(C_y,u_y)$, $K\!=\!\text{bin}(C_z,u_z)$; $A\!=\![I,J,K]_{\text{ax}}$, $U\!=\![I,J,K]_u$, $V\!=\![I,J,K]_v$
\State $O_{\text{fwd}}\leftarrow$ lexicographic order for $\mathrm{sgn}$; $\mathcal{O}\leftarrow[O_{\text{fwd}}]$ (optionally add reverse); init $\deg\!=\!0$, $\mathcal{E}\!=\!\emptyset$, $\mathcal{S}\!=\!\emptyset$; prebuild KD-trees if gates enabled; set $s,r$ from $s_{\text{vox}},r_{\text{vox}}$
\For{$O\in\mathcal{O}$}\For{$i\in O$} \If{$\deg[i]\!=\!\text{deg}_{\max}$} \textbf{continue} \EndIf
  \State $\mathcal{C}\leftarrow\emptyset$; \For{$j=1..N,\ j\neq i$}
    \State $dA\!=\!A[j]\!-\!A[i]$; \If{$\mathrm{sgn}\,dA\le0$ or $|dA|\!>\!A_{\max}$ or $|U[j]\!-\!U[i]|\!+\!|V[j]\!-\!V[i]|\!>\!R_{\text{side}}$} \textbf{continue} \EndIf
    \State Push $(j,\Delta_{\text{ax}}(i,j),\text{lat}(i,j))$ to $\mathcal{C}$
  \EndFor
  \State Sort $\mathcal{C}$ by $(\Delta_{\text{ax}}\!\uparrow,\text{lat}\!\uparrow)$; keep first $K_{\mathrm{cand}}$
  \For{$(j,\cdot,\cdot)\in\mathcal{C}$}
    \If{$\deg[i]\!=\!\text{deg}_{\max}$ or $\deg[j]\!=\!\text{deg}_{\max}$} \textbf{continue} \EndIf
    \State $e\!=\!(\min(i,j),\max(i,j))$; \If{$e\in\mathcal{S}$} \textbf{continue} \EndIf
    \If{\textsc{Cluster} on} reject if $\min\{d(\mathrm{KD}_{L[j]},C[i]),d(\mathrm{KD}_{L[i]},C[j])\}\!>\!\tau$ \EndIf
    \If{\textsc{Between} on} let $\phi\!=\!\textsc{HitFrac}(C[i],C[j];\mathrm{KD}_\cup,s,r)$; \If{$\phi\!<\!\phi_{\min}$} \textbf{continue} \EndIf \EndIf
    \State Append $(i,j,\|C[j]\!-\!C[i]\|_2)$ to $\mathcal{E}$; insert $e$ into $\mathcal{S}$; $\deg[i]\!+\!=\!1$; $\deg[j]\!+\!=\!1$
  \EndFor
\EndFor\EndFor
\State \textbf{return} $\mathcal{E},\deg$
\end{algorithmic}
\label{algo:detGraph}
\end{algorithm}

\section{Results and Discussion\label{sec:results}} \addvspace{10pt}
The first-subsection discusses results for two manufactured problems and the latter subsection measures and approximates a 3D detonation lattice using the said graph algorithm.

\subsection{Generated lattice\label{sec:results_toy}} \addvspace{10pt}
The two sub-sections present analysis of two different types of feature important in the lattice structure, first is the cellular volume and second is the bounding structure for this enclosed volume. 

\subsubsection{Volumetric Dataset}
\begin{figure}[!ht]
    \centering
  \includegraphics[width=0.85\linewidth]{ 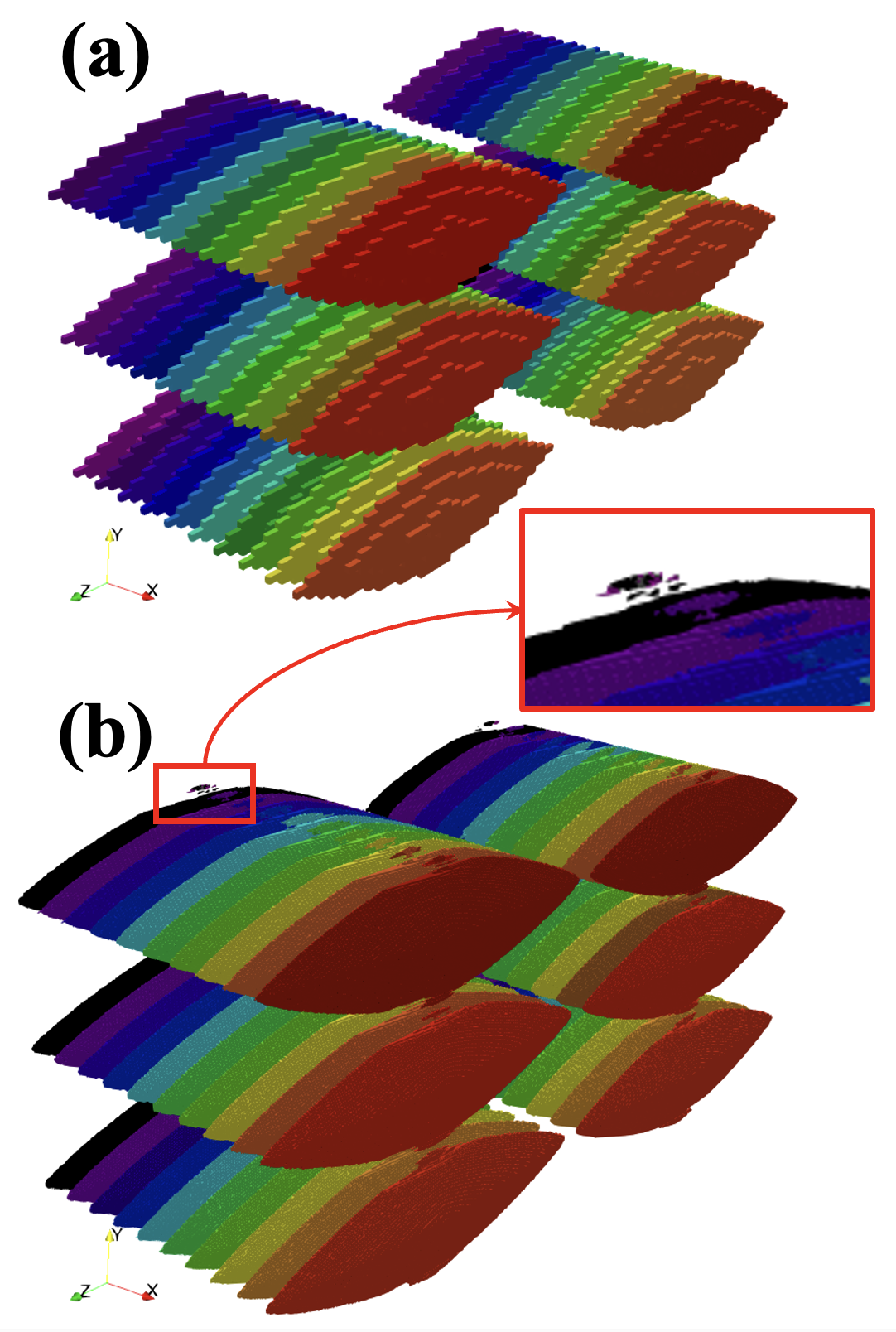}
  \caption{\footnotesize Plot for 3-D generated lattice segmented by the model: (a) $n_x$=60 and (b) $n_x$=480. Red box shows mask fragmentation due to over-segmentation in $n_x$=480 case.}
  \label{fig:genlat}
\end{figure}

To assess volumetric segmentation and detection efficacy, a synthetic 3D dataset was constructed using a fixed geometric arrangement of tightly packed, ellipsoid-shaped cell volumes, as shown in Fig.~\ref{fig:genlat}. \textcolor{black}{This synthetic lattice is not the raw numerical soot foil. The numerical soot foil is obtained separately by recording the maximum pressure experienced by each computational cell over the selected simulation duration, yielding a three-dimensional scalar field $p_{max}$. The high-pressure soot-foil boundary/envelope is then used to define cell-like volumetric regions. The synthetic dataset in Fig.~\ref{fig:genlat} is designed to mimic such enclosed detonation-cell volumes under controlled conditions.} The objects are arranged on a regular lattice with minimal inter-object spacing to induce frequent contacts and occlusions, thereby increasing the difficulty of segmentation and detection.
Experiments vary only the input sampling to the model, the ``cell count'' per axis of the volumetric grid ($n_x$), while keeping the underlying geometry unchanged (base geometry defined at $n_x$=240). Input resolutions range from 60 to 480 voxels per axis (Fig.~\ref{fig:genlat}), providing a controlled progression from coarse to refined sampling to quantify how much structure is recoverable as discretization improves.

\begin{table}[ht]
\centering
\caption{\footnotesize Effect of cell count ($n_x$) on prediction error (\%) (top row), the size of the lattice file (middle row) and number of masks generated by the model (bottom row).}
\label{tab:error}
\resizebox{\columnwidth}{!}{\begin{tabular}{cccc}\hline
\textbf{060}
& \textbf{120}
& \textbf{240}
& \textbf{480}
\\\hline


14.23 $\pm$ 2.37 &  14.62 $\pm$ 1.31     & 02.38 $\pm$ 0.19    & 00.73 $\pm$ 0.49 \\
0.66 MB &  5.20 MB   & 41.50 MB    & 331.90 MB \\
60 & 60 & 60 & 1101 \\
\hline
\end{tabular}}
\end{table}

From Table~\ref{tab:error}, prediction error remains high at coarse samplings (14.23\% at 60; 14.62\% at 120, see Fig.~\ref{fig:genlat}(a)), then drops sharply near the native geometric scale (2.38\% at 240) and reaches its lowest value at 480 (0.73\%). File size grows approximately cubically with resolution—0.66 MB to 331.90 MB—reflecting the volumetric scaling law. Notably, mask proliferation is negligible until 480 cells per axis (60 masks at 60 to 240 versus 1101 at 480), indicating over-segmentation emerges only at the finest input sampling. Overall, sampling near the base geometry ($n_x$=240) captures structure accurately with modest storage, while 480 offers marginal error gains at substantially higher storage and a risk of excessive mask fragmentation (see Fig.~\ref{fig:genlat}(b)). \textcolor{black}{Therefore, in practice, the preferred resolution should approximately match the native refinement scale of the generated lattice data, or the simulation grid (if data is provided on the same grid).}

\subsubsection{Skeletal Dataset\label{sec:results_toy2}}

\begin{figure*}[!h]
    \centering
  \includegraphics[width=0.95\linewidth]{ 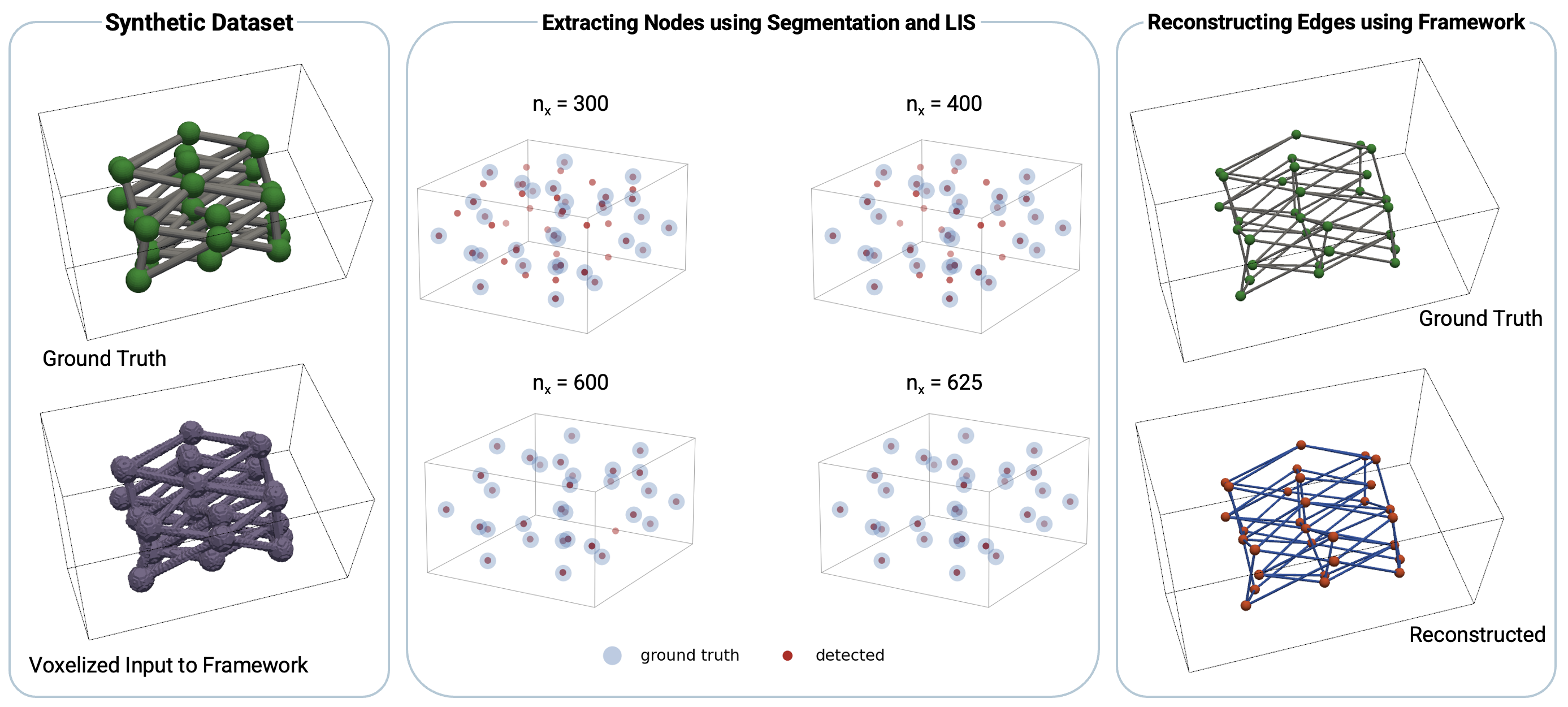}
    \caption{\footnotesize Synthetic 3D lattice benchmark used to evaluate node detection and edge reconstruction. The left panel shows the prescribed ground-truth graph and the corresponding voxelized input provided to the framework. The middle panel compares ground-truth node locations with detected nodes obtained from segmentation and LIS localization for different segmentation thresholds. The right panel compares the prescribed ground-truth connectivity with the reconstructed graph. The known graph contains 30 nodes and 64 edges, enabling quantitative assessment of reconstruction accuracy and parameter sensitivity.}
  \label{fig:data2}
\end{figure*}

\begin{figure}[!h]
    \centering
  \includegraphics[width=0.85\linewidth]{ 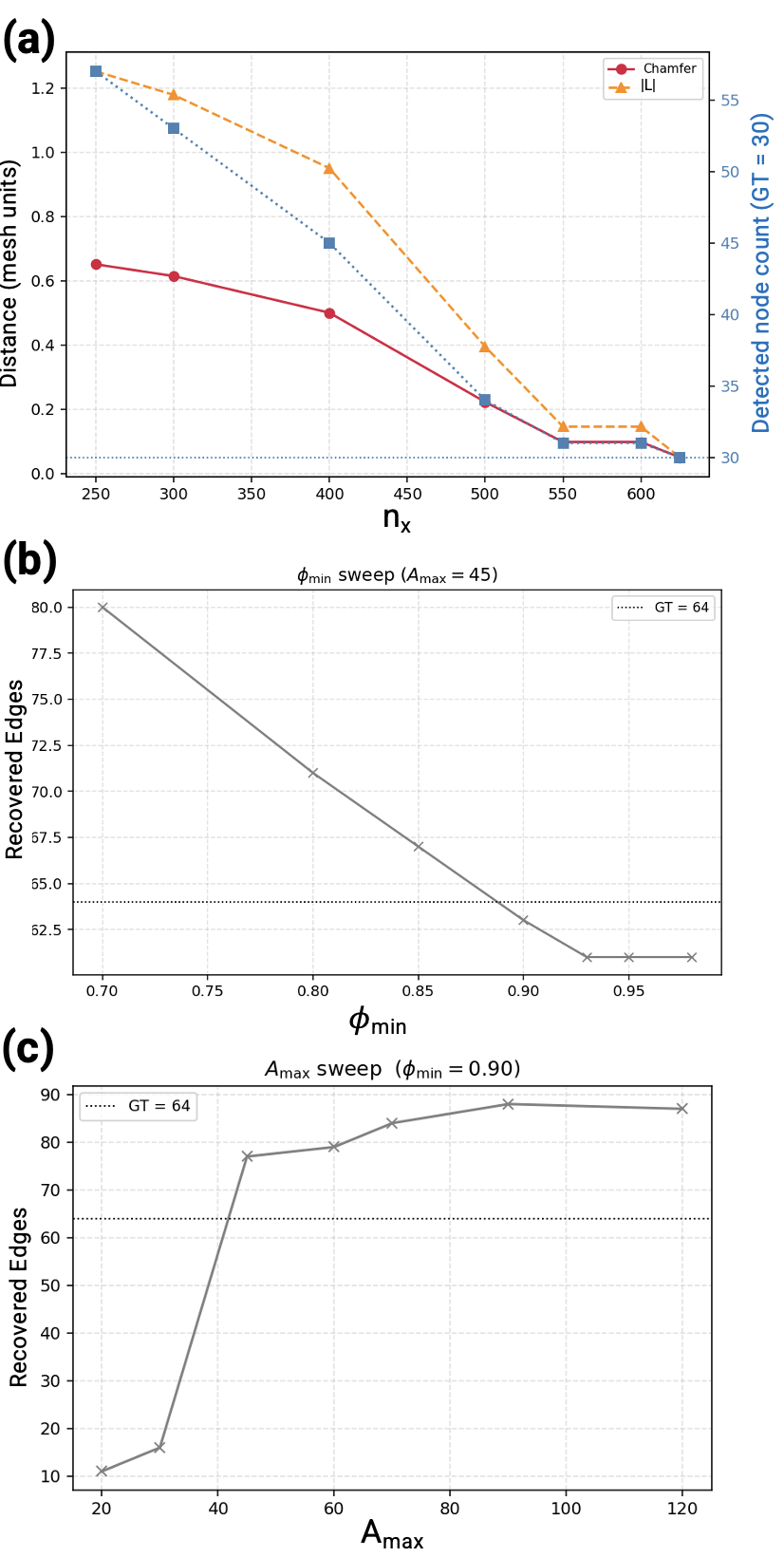}
    \caption{\footnotesize Sensitivity of the lattice-reconstruction workflow to the principal algorithmic parameters. (a) Effect of the segmentation size threshold on node detection, quantified using Chamfer distance, localization error, and detected-node count (b) Effect of the occupancy-continuity threshold $\phi_{\min}$ on recovered edge count at fixed $A_{\max}=45$. (c) Effect of the axial candidate-search window $A_{\max}$ at fixed $\phi_{\min}=0.90$. The dashed reference line in (b,c) denotes the ground-truth edge count, $E_{\mathrm{GT}}=64$.}
  \label{fig:parasweep_2}
\end{figure}

\textcolor{black}{
To quantify the sensitivity of the proposed lattice-reconstruction workflow, a synthetic three-dimensional soot-foil lattice (skeletal dataset) with known graph connectivity was used as a benchmark, as shown in Fig.~\ref{fig:data2}(left). The prescribed ground-truth graph contains 30 nodes and 64 edges. The synthetic geometry was first voxelized and then processed using the same two-stage workflow used for the detonation-lattice analysis: node extraction by segmentation followed by LIS localization, and edge reconstruction using the novel occupancy-based graph framework. This benchmark enables direct assessment of the principal algorithmic inputs because both the node locations and edge connectivity are known a-priori.
The first sensitivity parameter is $n_x$, that sets the minimum mask size used during the segmentation stage, affects node detection stage. This parameter removes segmented components containing fewer than $n_x$ voxels and therefore primarily controls over-segmentation of small fragments. Since $n_x$ is a voxel-count threshold, its physical meaning depends on the voxel volume. For a grid spacing $\Delta x$, $\Delta y$, and $\Delta z$, the corresponding physical volume threshold is $V_{\mathrm{thr}} \approx n_x \Delta x \Delta y \Delta z$. Thus, $n_x$ should be interpreted as a resolution-dependent representation of a physical size cutoff rather than as an intrinsic geometric quantity, similar to $\S$~\ref{sec:results_toy}. As shown in Fig.~\ref{fig:parasweep_2}(a), small values of $n_x$ retain many spurious fragmented masks, producing 57 detected nodes at the lowest tested value, while the ground truth contains only 30 nodes. Increasing $n_x$ progressively suppresses these fragments and reduces the Chamfer distance \footnote{Here, \(L=\frac{1}{|D|}\sum_{d\in D}\min_{g\in G}\|d-g\|\) denotes the detected-set-to-ground-truth distance, \(L_{\mathrm{inv}}=\frac{1}{|G|}\sum_{g\in G}\min_{d\in D}\|g-d\|\) denotes the inverse ground-truth-to-detected-set distance, and \(d_{\mathrm{Chamfer}}=\frac{1}{2}(L+L_{\mathrm{inv}})\) is the symmetric Chamfer distance.} between the detected and ground-truth node sets. This is also shown in Fig.~\ref{fig:data2}(middle), where predicted nodes (red) are overestimated compared to ground-truth nodes. At $n_x=625$, the method recovers the correct number of nodes, $N_{\mathrm{det}}=30$, with a residual Chamfer distance of approximately 0.05 mesh units, indicating that the remaining discrepancy is at the sub-voxel localization level. 
}
\begin{table}[ht]
\centering
\caption{\footnotesize Sensitivity of the reconstructed edge count to the edge-acceptance parameters. The ground-truth synthetic lattice contains $E_{\mathrm{GT}}=64$ edges. The error is reported as $\Delta E(\%)=100(E_{\mathrm{rec}}-E_{\mathrm{GT}})/E_{\mathrm{GT}}$, where positive values indicate over-reconstruction and negative values indicate under-reconstruction. For $\phi_{\min}$ sweep, $A_{\max}=45$ was held constant and for $A_{\max}$ sweep, $\phi_{\min}=0.90$ was set constant.}
\label{tab:edge_sensitivity}
\footnotesize
\[
\begin{array}{ccc|ccc}
\hline
\multicolumn{3}{c|}{\phi_{\min}\ \mathrm{sweep}} &
\multicolumn{3}{c}{A_{\max}\ \mathrm{sweep}} \\
\hline
\phi_{\min} & E_{\mathrm{rec}} & \Delta E\ (\%) &
A_{\max} & E_{\mathrm{rec}} & \Delta E\ (\%) \\
\hline
0.70 & 80 & +25.0 & 20  & 11 & -82.8 \\
0.80 & 71 & +10.9 & 30  & 16 & -75.0 \\
0.85 & 67 & +4.7  & \textbf{45}  & 77 & +20.3 \\
\textbf{0.90} & 63 & -1.6  & 60  & 79 & +23.4 \\
0.93 & 61 & -4.7  & 70  & 84 & +31.3 \\
0.95 & 61 & -4.7  & 90  & 88 & +37.5 \\
0.98 & 61 & -4.7  & 120 & 87 & +35.9 \\
\hline
\end{array}
\]
\end{table}

\textcolor{black}{
Next, two sensitivity parameters control edge formation. For each candidate node pair, the between-node occupancy fraction $\phi$ measures the fraction of sampled points along the inter-node segment that lie within the occupied wave-track material. An edge is accepted only if $\phi \geq \phi_{\min}$. The parameter $\phi_{\min}$ therefore controls the strictness of the material-continuity criterion. As shown in Fig.~\ref{fig:parasweep_2}(b), decreasing $\phi_{\min}$ admits excessive false connections: at $\phi_{\min}=0.70$, the algorithm recovers 80 edges compared with the 64 ground-truth edges. Increasing $\phi_{\min}$ reduces these over-connections by requiring stronger occupancy continuity along the candidate path. However, overly strict values begin to reject legitimate edges whose occupancy is imperfect because of voxelization, local curvature, or segmentation roughness. In the present synthetic test, values near $\phi_{\min}=0.90$ give the closest agreement with the ground-truth edge count before the additional graph-consistency guards are applied.
}
\textcolor{black}{
The axial search parameter $A_{\max}$ limits the maximum allowable separation between candidate node pairs along the search direction. This parameter controls the candidate set before the occupancy criterion is evaluated. If $A_{\max}$ is too small, valid long edges are never proposed, producing under-reconstruction. This behavior is evident in Fig.~\ref{fig:parasweep_2}(c), where $A_{\max}=20$ and $A_{\max}=30$ recover only 11 and 16 edges, respectively. Once the search window is sufficiently large, the true neighboring nodes are included in the candidate set. However, increasing $A_{\max}$ further also admits distant non-neighboring nodes, which can generate skip connections or other false edges if the inter-node path passes through occupied material. Thus, $A_{\max}$ exhibits the expected trade-off between candidate starvation at small values and spurious long-range connectivity at large values.
}
\textcolor{black}{
To make the parameter sensitivity quantitative, Table~\ref{tab:edge_sensitivity} reports the recovered edge count and its signed percent deviation from the known ground truth, $E_{\mathrm{GT}}=64$. The $\phi_{\min}$ sweep shows the expected transition from over-reconstruction to mild under-reconstruction as the occupancy-continuity requirement is tightened. At low $\phi_{\min}$, non-adjacent node pairs can be falsely connected because the candidate path intersects occupied lattice material, producing a $+25.0\%$ edge-count error at $\phi_{\min}=0.70$. Increasing $\phi_{\min}$ suppresses these false connections, with the edge count approaching the ground truth near $\phi_{\min}=0.85$--$0.90$. In contrast, the $A_{\max}$ sweep shows that too small an axial search window starves the candidate set, giving severe under-reconstruction at $A_{\max}=20$ and 30. Once $A_{\max}$ is sufficiently large to include true neighboring nodes, the recovered count exceeds the ground truth because distant non-neighboring pairs also enter the candidate set. These results provide a quantitative operating range for the two principal edge-reconstruction parameters and identify the failure modes associated with under- and over-selection of candidate connections.
}
\textcolor{black}{
The skeletal synthetic benchmark finally uses $n_x=625$, $\phi_{\min}=0.95$, $A_{\max}=50$ and a minimum edge length $\ell_{\min}=3.0$. With these parameters, the reconstructed graph contains 30 nodes and 64 edges, matching the prescribed ground truth, as shown in Fig.~\ref{fig:data2}(right). The parameter sweeps in Fig.~\ref{fig:parasweep_2} therefore show that the method is not tuned to a single isolated parameter value: node recovery remains stable once fragment masks are removed, while edge recovery is governed by physically interpretable limits on material continuity, axial search extent, and intermediate-node exclusion.
}

\begin{figure*}[!ht]
    \centering
  \includegraphics[width=0.95\linewidth]{ 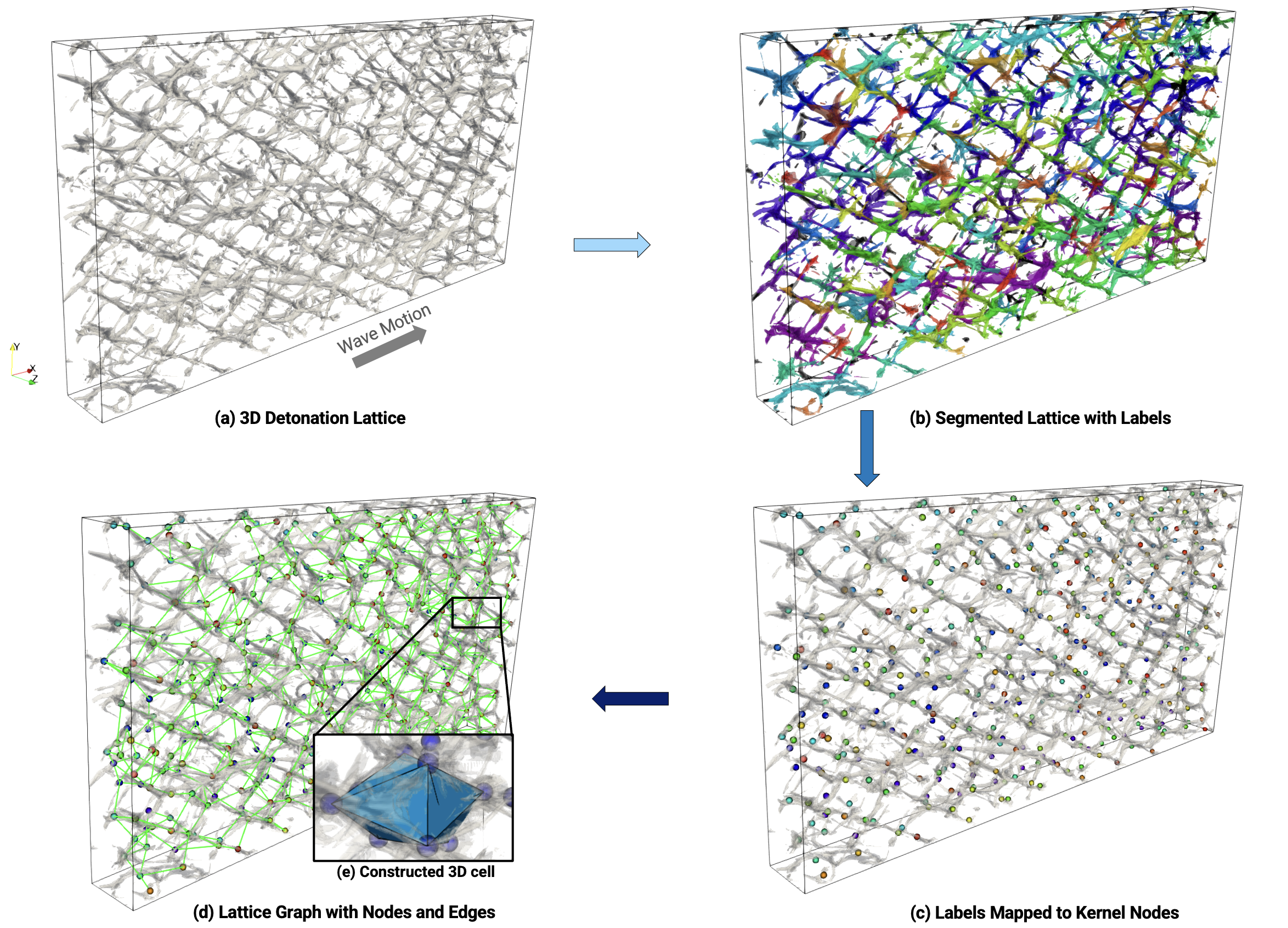}
  \caption{\footnotesize Overall workflow of the segmentation and graph construction procedures on a 3D numerical detonation lattice.}
  \label{fig:Graph}
\end{figure*}

\subsection{3D Detonation simulation based reconstruction\label{sec:results-3Dsim}} \addvspace{10pt}

The algorithm is evaluated on 3D lattice field generated from a gaseous detonation simulation of \textcolor{black}{a confined stoichiometric ethylene–air mixture at a pre-detonation temperature of 500 K and pressure of 1 atm. The analyzed domain is the long rectangular configuration (Case 4) with $L_x=1453\times{l_{\mathrm{ind}}}$, $L_y=78\times{l_{\mathrm{ind}}}$, and $L_z=15 \times {l_{\mathrm{ind}}}$ (aspect ratio $5.33:1$), initialized with 11 detonation kernels (each kernel is composed of $\mathrm{N}_2$ at $P$ = 65 atm and $T$ = 2800 K) and all walls satisfy no-slip boundary conditions. The one-dimensional induction length for this mixture is $l_{ind}$ = 516 $\mu$m.} The finest grid spacing is 9.766 \(\mu\)m, which yields about 52 cells per induction length resulting in 800 million cells. This resolution is sufficient to resolve the cellular detonation structure, including triple points, Mach stems, and collision zones, ensuring a sufficiently reliable representation of the underlying physics. 

\textcolor{black}{The 3D simulation was advanced sufficiently long (identified by the stabilization of both the mean wavefront velocity and the mean pressure profile evaluated on a z-plane slice at the y-midplane~\cite{Abiola2026_isoc}) before extracting the lattice statistics, so that the initial transient was removed and a statistically established cellular pattern was obtained. The resulting structure is therefore best characterized as near-regular rather than strongly irregular or transitional. Nevertheless, the cells are not perfectly periodic: local boundary distortions arise from transverse-wave interactions, variations in triple-point trajectories, confinement, and geometric effects along the propagation direction. Accordingly, the reported geometric quantities should be interpreted as statistical descriptors of this specific near-regular confined premixed detonation case. More irregular or unstable detonation regimes may exhibit broader distributions of cell volume, surface area, and connectivity.}
Additional details are discussed in~\cite{Abiola2026_isoc}.

Following the workflow in Fig.~\ref{fig:Graph}, the detonation lattice is first extracted by tracking the maximum pressure observed by each cell in the numerical domain (plot (a)). This data is stored as a point-based structured grid which is then sliced along the transverse direction for the segmentation model to process. The output of the SAM model assigns a unique ID to a cluster of cells as discussed in $\S$\ref{sec:results_toy} (plot (b)). To construct the graph, each cell cluster is mapped to a unique point that is both interior and boundary-aware (plot (c)). This step is essential as it determines the collision sites in the lattice that define the shape’s boundary points. The edges are subsequently constructed between these nodes (plot (d)) using the algorithm in $\S$\ref{sec:GraphCon} to obtain the 3D cell (plot(e)).  
All physical quantities are reported in the mesh’s native coordinate units; no unit conversion was applied. The measurements are computed directly from the geometry of the closed surface (vertex positions and triangulated faces), not from a voxel/grid discretization, i.e., no voxels are used. \textcolor{black}{In the present implementation, the reconstructed cell surfaces are therefore represented as triangulated meshes, and all reported geometric quantities are evaluated on triangular facets rather than quadrilateral or general polygonal faces. If a larger polygonal surface patch arises during reconstruction, it is represented through triangulation, which does not affect the downstream geometric analysis because all measurements are performed on the triangulated surface geometry.}

\begin{figure}[!ht]
    \centering
  \includegraphics[width=0.85\linewidth]{ 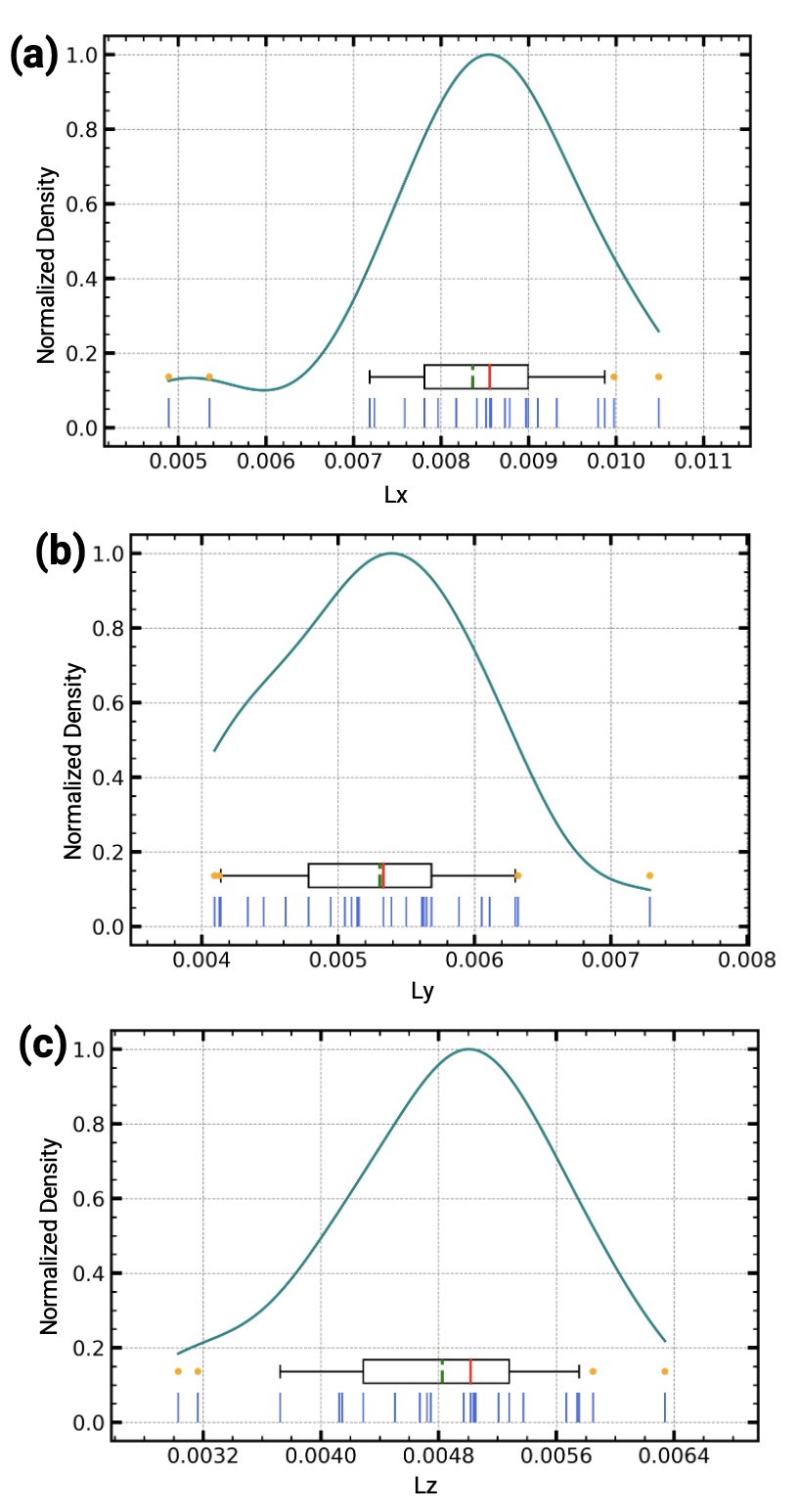}
  \caption{Plot of geometric features from 25 samples: (a) major-axis length, (b) minor-axis length, each shown as a KDE (normalized) with stick marks indicating individual  samples (blue) and an inline boxplot (median = red line, mean = green dashed line, whiskers 5$^{\text{th}}$–95$^{\text{th}}$ percentiles); (c) joint density of volume and aspect ratio (major/minor axis) with overlaid points (normalized density color scale).}
  \label{fig:stats1}
\end{figure}
Fig.~\ref{fig:stats1} shows the one-dimensional distributions (from 25 samples) of the lengths of the major and minor axes (mesh units) in plots (a), (b) and \textcolor{black}{(c)}: the teal curve is a kernel density estimate (KDE) (normalized with peak value); all features appear unimodal. The cell geometry is systematically elongated in the X-axis with moderate variability in length scales but substantial variability in volume. The distribution of the major axis (L$_x$) centers near 0.00839 (median), with a slightly larger mean (0.00851) and a fairly narrow spread in the middle. The negative skew is evident with a longer left tail: few smaller samples deviate trend to the low-end. However, most samples lie in a narrow band around 0.008–0.009. In contrast, minor axis (L$_y$) distribution is more concentrated around its center (median = 0.00403) and only slightly right-skewed: the mean (0.00403) essentially matches the median, with a few modestly larger values. A similar trend follows for L$_z$ distributions with a median around 0.005 slightly larger than the mean $\simeq$ 0.0048. Table~\ref{tab:stats1} reports different statical measures (from Fig.~\ref{fig:stats1}) for cell lengths ($L$) along each axis and the geometric volume of the 3D cell. The coefficients of variation ($\sigma$/$\mu$) are modest for lengths, approx. 15-17\% across the three $L$s. In contrast, volume is highly dispersed, indicating a typical pattern when small differences in linear size propagate cubically into volume. For all three lengths, $M > \mu$, suggesting a slight right-skew (few larger values). Volume shows the opposite trend, indicating a left-skewed distribution, consistent with multiplicative variation in size.

\begin{table}[ht]
\centering
\caption{\footnotesize Summary statistics: the mean ($\mu$), standard deviation ($\sigma$), and median ($M$) of the extents $L_x$, $L_y$, $L_z$ and V (volume) for 25 samples in native units of the mesh.}
\label{tab:stats1}
\resizebox{\columnwidth}{!}{\begin{tabular}{ccccc}\hline
\textbf{Measure}
& \textbf{$L_x$}
& \textbf{$L_y$}
& \textbf{$L_z$}
& \textbf{V}
\\\hline


$\mu$ & 8.36e-3 &  5.30e-3  & 4.82e-3  & 2.11e-7 \\
$\sigma$ & 12.82e-4 &  7.88e-4  & 8.09e-4 & 2.36e-7 \\
$M$ & 8.55e-3 & 5.33e-3 & 5.01e-3 & 1.33e-7 \\
\hline
\end{tabular}}
\end{table}

\begin{figure}[!ht]
    \centering
  \includegraphics[width=0.85\linewidth]{ 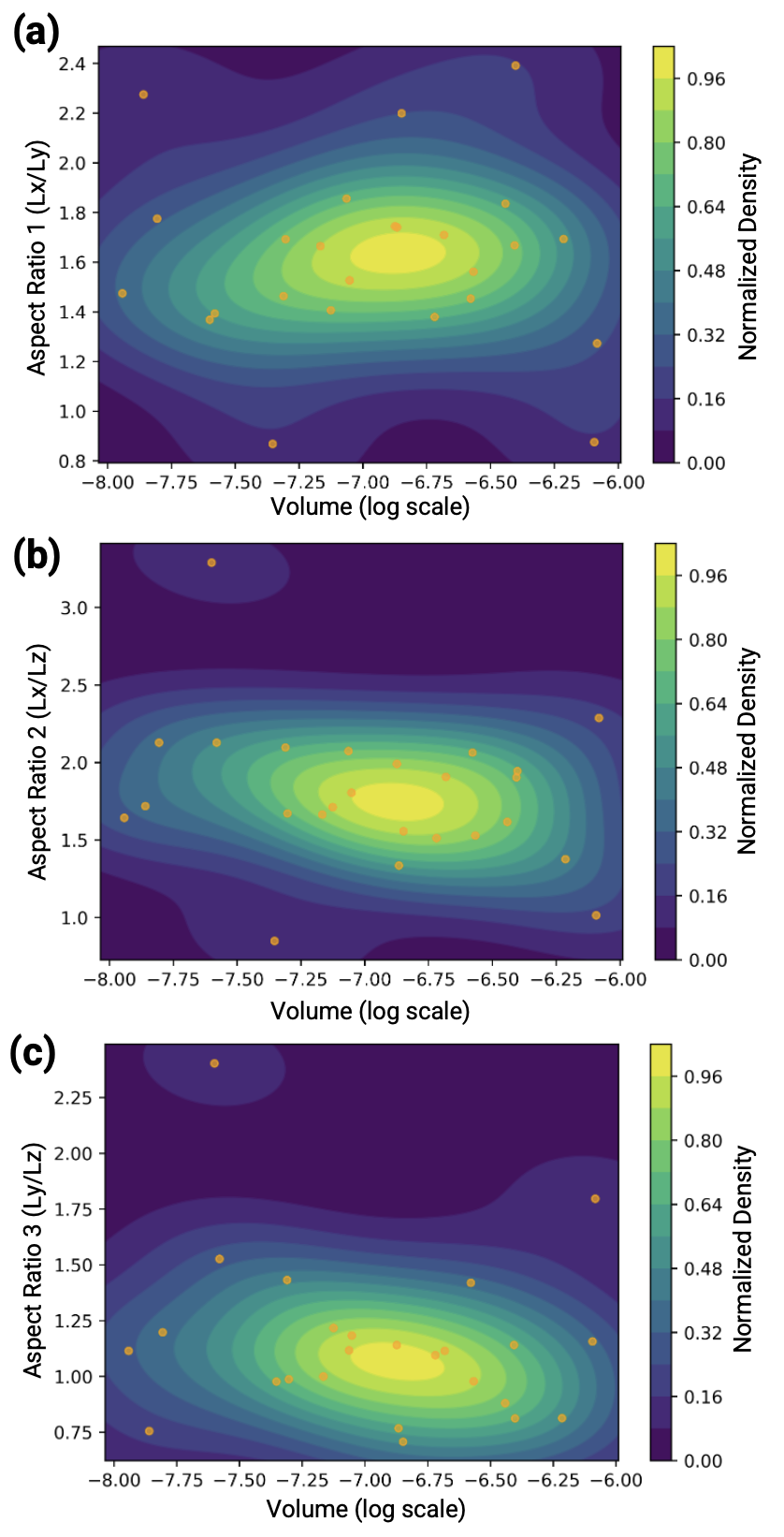}
  \caption{\footnotesize Plot of joint density of volume and aspect ratio (AR) from 25 samples with overlaid points (normalized density color scale): (a) AR$_1$=$L_x$/$L_y$, (b) AR$_2$=$L_x$/$L_z$ and (c) AR$_3$=$L_y$/$L_z$.}
  \label{fig:stats2}
\end{figure}

Figure~\ref{fig:stats2} shows the joint distribution of volume (log-scale) with different aspect ratios (AR), where contours represent normalized density and orange dots are the samples. Plots (a) and (b) show that most shapes cluster at intermediate volumes with moderate aspect ratios (1.6 to 2.2), with lighter contours (and scattered points) extending toward larger volumes and higher aspect ratios. This observation is consistent with the elongation of X-axis over Y and Z. The plot (c) for AR$_3$ is clustered close to 1.0–1.2, implying Y and Z extents are comparatively similar, as observed by Oladipo et al.~\cite{Abiola2026_isoc}. Across all three, the contours are broadly ellipsoidal with only a slight tilt, suggesting at most a weak dependence of aspect ratio on volume: larger volumes do not strongly skew the AR, though the highest densities sit at intermediate volumes and moderate anisotropy.


\section{Conclusion\label{sec:final}} \addvspace{10pt}

A graph-based multi-axis framework was developed to analyze and quantify the three-dimensional morphology of detonation cells. \textcolor{black}{Evaluated with two controlled lattice test cases that mimic detonation-cell volumes and its lattice structure, the workflow predicts cell volume within 2\% error, validating the segmentation phase of the method.} The framework was then applied to 3D numerical data to extract statistical measures for axial length scales $(L_x,L_y,L_z)$, cell volume $V$, and their joint distributions. The resulting length distributions are unimodal and indicate systematically elongated cells along the $x$-direction, with modest variability in the linear extents (coefficients of variation $\approx 15$--$17\%$). In contrast, $V$ exhibits substantially larger dispersion, consistent with cubic amplification of small variations in length. The joint-PDFs of $V$ and aspect ratios show clustering at intermediate volumes with moderate anisotropy ($\mathrm{AR}\approx 1.6$--$2.2$) and weak coupling between aspect ratio and volume, while the extents of $y$--$z$ remain comparatively similar. Overall, the method provides robust, physically interpretable 3D cell statistics suitable for subsequent studies of confined triple-point interactions and the growth of cellular instabilities.

\acknowledgement{CRediT authorship contribution statement} \addvspace{10pt}
{\bf VS}: designed research, developed software, performed research, analyzed data, wrote—original draft, review and editing.  
{\bf VR}: designed research, developed software, analyzed data, wrote—review and editing, project administration, funding acquisition.

\acknowledgement{Declaration of competing interest} \addvspace{10pt}
The authors declare that they have no known competing financial interests or personal relationships that could have appeared to influence the work reported in this paper.

\acknowledgement{Acknowledgments} \addvspace{10pt}
The authors acknowledge support from Air Force Office of Scientific Research (AFOSR) under Grant No. FA9550-24-1-0017, with Dr. Chiping Li as Program Officer. The authors thank Abiola Oladipo for providing 3D detonation data.

\footnotesize
\baselineskip 9pt

\clearpage
\thispagestyle{empty}
\bibliographystyle{proci}
\bibliography{PCI_LaTeX}


\newpage

\small
\baselineskip 10pt


\end{document}